\documentclass[letterpaper]{article} % DO NOT CHANGE THIS
\usepackage{aaai2026}  % DO NOT CHANGE THIS
\usepackage{times}  % DO NOT CHANGE THIS
\usepackage{helvet}  % DO NOT CHANGE THIS
\usepackage{courier}  % DO NOT CHANGE THIS
\usepackage[hyphens]{url}  % DO NOT CHANGE THIS
\usepackage{graphicx} % DO NOT CHANGE THIS
\usepackage{natbib}  % DO NOT CHANGE THIS AND DO NOT ADD ANY OPTIONS TO IT
\usepackage{caption} % DO NOT CHANGE THIS AND DO NOT ADD ANY OPTIONS TO IT
\usepackage{algorithm}
\usepackage{algorithmic}
\usepackage{amsmath} 
\usepackage{booktabs}   
\usepackage{multirow}   
\usepackage{siunitx}    
\usepackage{amsmath}   
\usepackage{amssymb}    
\usepackage{threeparttable}
\usepackage{booktabs}
\usepackage{graphicx}
\usepackage{extpfeil}

\usepackage{stmaryrd}
\usepackage{makecell}

\usepackage[table]{xcolor}
\definecolor{row_color}{HTML}{EFF7FF}
\definecolor{myblue}{HTML}{548dc0}
\definecolor{myred}{HTML}{ff4746}
\definecolor{mygreen}{HTML}{86ba63}

\def\eg{\emph{e.g.}}

\usepackage{newfloat}
\usepackage{listings}
\DeclareCaptionStyle{ruled}{labelfont=normalfont,labelsep=colon,strut=off} % DO NOT CHANGE THIS
\floatstyle{ruled}
\newfloat{listing}{tb}{lst}{}
\floatname{listing}{Listing}
\title{UniHOI: Unified Human-Object Interaction Understanding via \\Unified Token Space}
\author{
    Panqi Yang\textsuperscript{\rm 1}\equalcontrib, 
    Haodong Jing\textsuperscript{\rm 1}\equalcontrib,
    Nanning Zheng\textsuperscript{\rm 1}, 
    Yongqiang Ma\textsuperscript{\rm 1}\thanks{Corresponding author}
}
\affiliations{
    \textsuperscript{\rm 1}State Key Laboratory of Human-Machine Hybrid Augmented Intelligence,\\
    National Engineering Research Center of Visual Information and Applications,\\
    and Institute of Artificial Intelligence and Robotics, Xi'an Jiao Tong University\\
    \{yangpq,jinghd\}@stu.xjtu.edu.cn, \{nnzheng,musayq\}@xjtu.edu.cn
}

\usepackage{bibentry}
\begin{document}

\maketitle
% \twocolumn[{%
% \renewcommand \twocolumn[1][]{#1}%
% \maketitle
% %首页图，突出我们方法在交互生成方面的优异性能（也要注意检测方面），同时介绍一下雷达图中部分指标；
% \begin{center}
%   \centering
%   \includegraphics[width=\textwidth]{./fig/homepage_cmyk.pdf}
%    \captionof{figure}{UniHOI is the first to achieve unified modeling for the two inverse tasks of HOI detection and generation. Through a unified token space, our method enables generalizable interaction semantics understanding and cross-task knowledge sharing. UniHOI achieves state-of-the-art results on most metrics for both HOI detection and generation. Here, HICO-D refers to the \textit{Rare} metric of the \textit{Default} split in the HICO-DET~\cite{hico-det}, other abbreviations follow similarly.}
%    \label{fig:homepage}
% \end{center}%
% }]

\begin{figure*}[t]
  \centering
  \includegraphics[width=\textwidth]{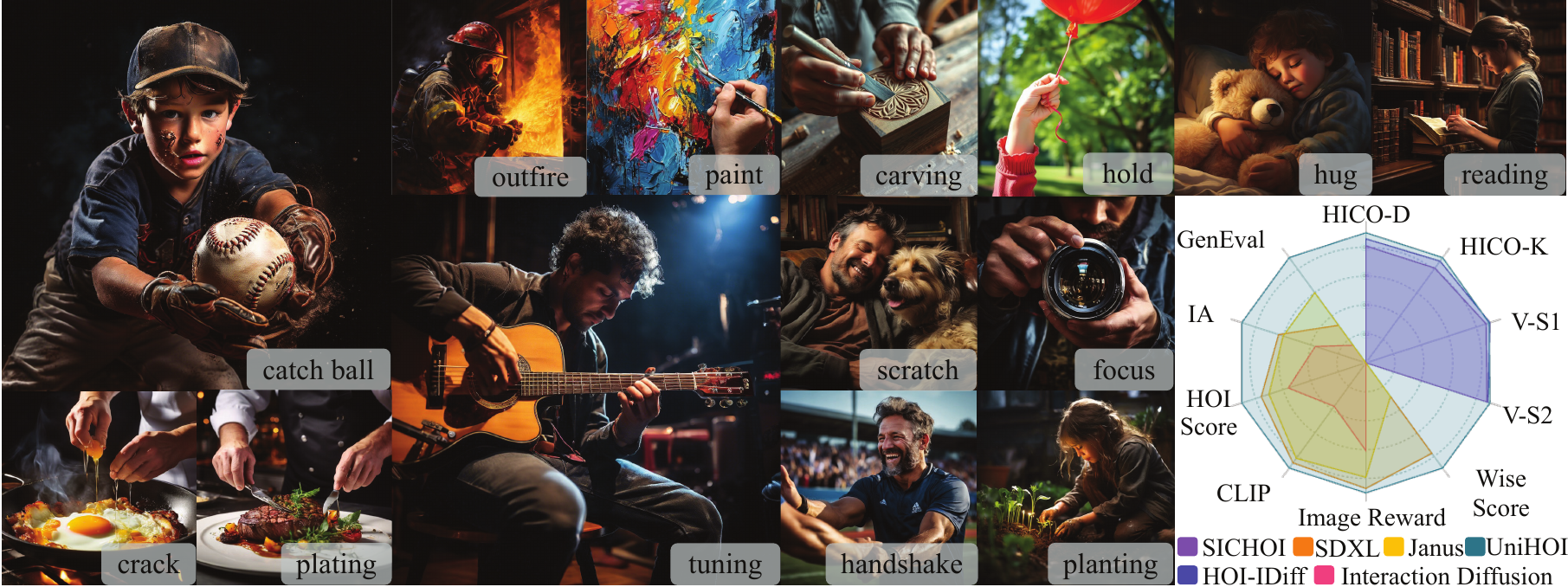}
  \caption{UniHOI is the first to achieve unified modeling for the two inverse tasks of HOI detection and generation. Through a unified token space, our method enables generalizable interaction semantics understanding and cross-task knowledge sharing. UniHOI achieves state-of-the-art results on most metrics for both HOI detection and generation. Here, HICO-D refers to the \textit{Rare} metric of the \textit{Default} split in the HICO-DET~\cite{hico-det}, other abbreviations follow similarly.}
  \label{fig:homepage}
\end{figure*}

\begin{abstract}
In the field of human-object interaction (HOI), detection and generation are two dual tasks that have traditionally been addressed separately, hindering the development of comprehensive interaction understanding. To address this, we propose UniHOI, which jointly models HOI detection and generation via a unified token space, thereby effectively promoting knowledge sharing and enhancing generalization.
Specifically, we introduce a symmetric interaction-aware attention module and a unified semi-supervised learning paradigm, enabling effective bidirectional mapping between images and interaction semantics even under limited annotations. Extensive experiments demonstrate that UniHOI achieves state-of-the-art performance in both HOI detection and generation. Specifically, UniHOI improves accuracy by 4.9\% on long-tailed HOI detection and boosts interaction metrics by 42.0\% on open-vocabulary generation tasks.
\end{abstract}


\section{Introduction}
\label{sec:intro}

%Background:讲明白问题的定义与背景，以及需要解决的问题
%这两种互逆任务的统一，想说明更高层次的交互理解；在解决原来各自任务问题的同时拔高一个层次。
The Human-Object Interaction (HOI) understanding encompasses both HOI detection (identifying $\langle \text{human}, \text{action}, \text{object} \rangle$ triplets in images) and HOI generation (synthesizing images conditioned on specified interactions). These two tasks are essentially inverse to each other, sharing highly related underlying semantic representations and reasoning processes. However, existing approaches largely treat detection and generation as isolated problems: most HOI generation methods~\cite{GLIGENOG,InteractDiffusion,interactgan} rely on explicit spatial constraints (\eg, bounding boxes), limiting their ability to generalize to novel or complex interactions. Meanwhile, state-of-the-art HOI detection models~\cite{hoi_det1,hoi_det3,hoi_det5} focus solely on recognition and remain decoupled from generation, which hinders knowledge sharing and demands extensive fine-grained annotations for each task. This significantly limits the scalability of HOI understanding in open-world scenarios.

%提出我们的想法：用一个统一的框架来完成hoi understanding
This motivates a key question: \emph{Can we unify HOI detection and generation within a single framework to fully exploit their shared semantic representations?} We contend that such unification would promote richer interaction understanding, improve data and knowledge efficiency, and enable cross-task and open-world generalization, thus paving the way for more holistic HOI reasoning.

%如何实现这个统一的框架呢？
%通过近期的mllm工作引出我们的核心insight，提供理论支撑
The recent emergence of Multimodal Large Language Models (MLLMs) offers a promising approach for unifying perception and generation within a single, expressive framework. Inspired by works like MMaDA~\cite{mmada} and particularly Liquid~\cite{liquid}, which demonstrated the potential of unified token spaces for joint vision-language modeling, we argue that a unified representation for interaction semantics holds the key to bridging the HOI detection-generation gap. To this end, our approach is grounded in two key insights:

%说明我们的insight:用简单易懂的话说明这个想法对这个任务的帮助
First, \textbf{Unified Token Space}: the semantic essence of a HOI triplet $\langle \text{human}, \text{action}, \text{object} \rangle$ can be effectively represented within the same discrete vocabulary as visual tokens derived from images (\eg, VQGAN~\cite{vqgan}). This enables bidirectional mapping within a MLLM, where a textual prompt like ``person feeding cat'' can evoke spatial relationships (``hand-bowl-cat'') during generation, while visual tokens can be decoded back into semantic triplets during detection. 

Second, \textbf{Dual Complementarity}: HOI detection and generation are inherently complementary; detection provides explicit interaction priors (\eg, typical spatial distributions and fine-grained annotations) to guide generation, while generation encourages the model to acquire richer, more compositional interaction representations. This reciprocal relationship improves both the expressiveness and generalization of the learned HOI representations, allowing the model to effectively acquire unified HOI semantic representations.

%引出我们的方法
Therefore, we propose \textbf{UniHOI}, the first unified semi-supervised MLLM for HOI detection and generation. UniHOI operates on a massively expanded vocabulary merging visual codebook tokens and text tokens, accepting either images (for detection) or text prompts (including structured HOI triplets for generation) to produce corresponding outputs. Specifically, our approach incorporates: (1) an \textbf{Interaction-Aware Attention (IAA)} module that injects HOI triplet embeddings into cross-attention layers to focus on interaction-relevant regions; (2) a \textbf{semi-supervised learning framework} grounded in cycle consistency within the discrete token space, which enables effective joint training of HOI detection and generation even under limited or heterogeneous supervision. By integrating strong supervision (HICO-DET~\cite{hico-det}), weak supervision (image-text pairs from LAION-SG~\cite{laion-sg}), and unlabeled images, this framework mitigates reliance on exhaustive HOI annotations and facilitates the learning of more generalizable interaction representations for both tasks. And our contributions are as follows:

\begin{itemize}
    \item We demonstrate that interaction-aware semantic representations can be jointly encoded and reasoned about in a unified discrete token space. Our \textbf{modality-aware unified token space} enables bidirectional mapping and compositional reasoning between HOI detection and image generation within a single model, surpassing traditional embedding-level alignment approaches.

     \item We propose an \textbf{Interaction-Aware Attention (IAA)} module with parameter-shared, symmetric cross-attention that explicitly encodes structured HOI semantics as relational priors for both detection and generation. This unified mechanism enables interpretable, context-aware cross-modal reasoning.

    \item We present a unified \textbf{semi-supervised learning strategy} based on cycle consistency in the shared token space, allowing effective training with mixed supervision and reducing annotation cost for open-world HOI recognition and generation. 

    \item Extensive experiments demonstrate that UniHOI achieves highly competitive performance on standard HOI detection and generation benchmarks. Specifically, UniHOI improves accuracy by 4.9\% on long-tailed HOI detection and boosts interaction metrics by 42\% on open-vocabulary generation tasks.

\end{itemize}

\section{Related Work}
\label{sec:relat}

%这段讲为什么要用MLLM来完成HOI Understanding 以及IAA的思路来源
\subsection{\textbf{Text-to-Image Models}}
Diffusion-based T2I models~\cite{GLIDE,Imagen,stablediffusion,dalle2} iteratively denoise latent text-conditioned embeddings~\cite{clip,t5}, generating high-resolution images~\cite{stablediffusion,sdxl,Imagen}. Recent studies~\cite{densediffusion,freecontrol,p2p,p2p-hrv} reveal that their intermediate representations encode rich semantics, supporting advanced image editing via feature-text interactions. Meanwhile, MLLM-based T2I models~\cite{liquid,mmada,janus-pro,showo} leverage large language models for enhanced multimodal understanding and unified vision-language reasoning and generation. For instance, MMaDA~\cite{mmada} adopts a unified diffusion framework for joint inference and generation, while Liquid~\cite{liquid} shows mutual benefits between understanding and generation within MLLM architectures. Based on these works, we explore how interaction-aware cross attention in MLLMs perceives and generates interaction semantics.

%模型整体框架图
\begin{figure*}[t]
  \centering
  \includegraphics[width=\textwidth]{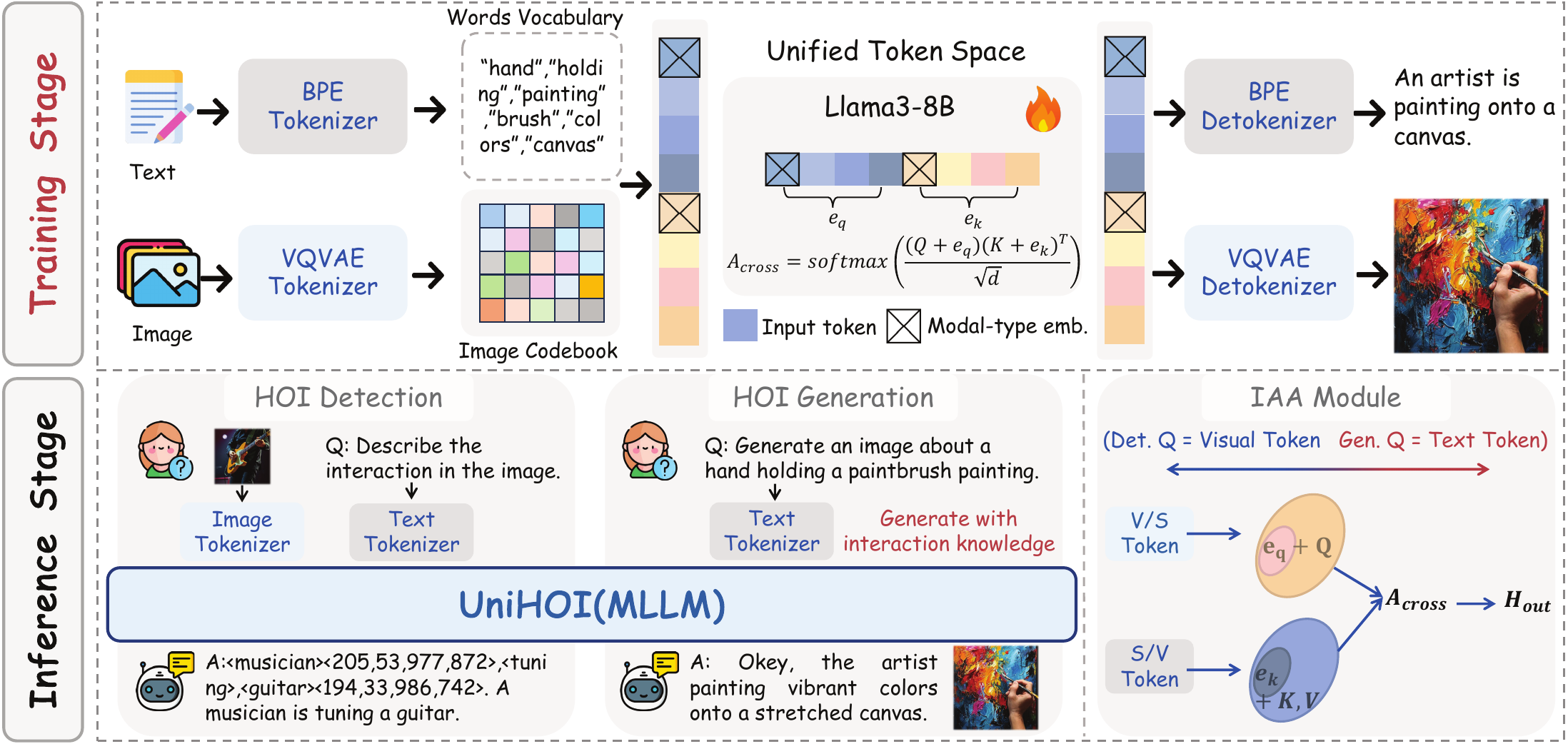}
   \caption{An overview of the UniHOI pipeline. The bottom-right shows the details of the IAA module, illustrating the bidirectional transformation between text tokens and visual tokens.}
   \label{fig:framework}
\end{figure*}

%这段讲一下HOI Understanding两个任务的背景以及为什么要把两个任务放在一起做，以及为什么我们要采用non-layout的方式来进行交互细节生成
\subsection{\textbf{HOI Understanding}}
HOI understanding encompasses two core tasks: HOI detection and HOI image generation. HOI detection~\cite{fgahoi,RLIP,thid,hotr} seeks to localize humans and objects and classify their interactions in the form of triplets (\eg, person, play, skateboard). Despite notable progress, this task remains constrained by the limited availability of high-quality annotated data. As the inverse task of detection, HOI image generation aims to generate images depicting specified interactions. Early approaches, such as InteractGAN~\cite{interactgan}, utilize human poses and object references, while subsequent methods~\cite{interactgan,InteractDiffusion} often depend on predefined spatial layouts alongside textual cues. However, such layout- or reference-based dependencies fundamentally constrain scalability and diversity. The requirement for expensive annotation and limited semantic expressiveness make it difficult to model complex or long-tailed interactions and to capture nuanced intentions from natural language. In this work, we aim to bridge the gap between these two tasks within a unified framework, alleviating the limitations imposed by scarce fine-grained annotations on HOI understanding, while also reducing the reliance on layout constraints for HOI image generation.

\section{Method}
\label{sec:method}
%简要概括下method重要章节的内容和整体逻辑梳理
% In this section, we introduce a unified framework that integrates HOI detection and image generation within a single architecture. Our approach leverages a unified token space, interaction-aware attention, and a cycle-consistent semi-supervised training strategy to enable explicit and mutually beneficial cross-modal learning.

\subsection{Overall Framework}
%介绍下UniHOI整体架构的设计思路以及输入输出（bbox为强监督部分学习内容，是符合method内容的）
UniHOI is a unified MLLM that bridges HOI detection and interaction-aware image generation within a shared multimodal latent space. Built upon a sequence-to-sequence Transformer, our architecture employs a unified vocabulary that incorporates both discrete visual tokens (\eg, from VQGAN~\cite{vqgan}) and semantic tokens (\eg, from text captions or structured HOI triplets), enabling bidirectional vision-language modeling, shown in Figure~\ref{fig:framework}. 

As illustrated by the following unified formulation,
\begin{equation}
    \mathcal{I} \;\rightleftharpoons\; \big\{\, \mathcal{H},\, \mathcal{B},\, \mathcal{C} \,\big\} \;\rightleftharpoons\; \mathcal{T}
\end{equation}
where $\mathcal{I}$ is the input image, $\mathcal{H}$ denotes HOI triplets, $\mathcal{B}$ the corresponding bounding boxes, $\mathcal{C}$ the interaction caption, and $\mathcal{T}$ a free-form or structured semantic prompt. UniHOI unifies HOI detection (image-to-triplet/caption/box) and HOI image generation (caption-to-image) within a single model, enabling flexible and direct mapping between visual inputs and interaction semantics for mutually enhanced recognition and generation.

%核心创新点1：讲明白我们的unified space和普通多模态对齐的区别以及和liquid的区别；以及我们具体是怎么设计的；
\subsection{Modality-Aware Unified Token Space}\label{sec:unified_token_space}

We present a modality-aware unified token space that enables direct bidirectional transformation between visual and semantic modalities. Unlike prior multimodal alignment approaches~\cite{llava,blip2}, which confine alignment to the embedding level and only support cross-modal similarity comparison, our method unifies both modalities at the token level by constructing a shared, discrete vocabulary and a common embedding manifold.

Formally, we define the joint vocabulary as
\begin{equation}
    \mathcal{V} = \mathcal{V}_{\text{code}} \cup \mathcal{V}_{\text{sem}},\quad
    \mathbf{E} \in \mathbb{R}^{|\mathcal{V}| \times d},
\end{equation}
where $\mathcal{V}_{\text{code}}$ and $\mathcal{V}_{\text{sem}}$ denote the sets of visual tokens (\eg, from images) and semantic tokens (\eg, text or HOIO triplets), respectively. All tokens share a unified, trainable embedding matrix, ensuring joint representation and generation.

Crucially, for each token, we introduce a learnable \textbf{modality-type embedding} in addition to the shared token embedding, explicitly encoding modality information. The final input embedding for each token is the sum of its token embedding and modality-type embedding, allowing the Transformer to distinguish and leverage both sources effectively. This explicit type encoding helps the model disambiguate token distributions, prevents modality confusion, and enables controllable cross-modal generation, whereas methods like Liquid~\cite{liquid} simply merge vocabularies without explicit modality conditioning.

With this design, both visual and semantic token sequences can serve as input or output for the same Transformer model, supporting the following explicit bidirectional mapping:
\begin{equation}
    \underbrace{\mathbf{x}_v \in \mathcal{V}_{\text{code}}^L}_{\text{visual tokens}}
    \xleftrightarrow[\text{generation}]{\text{detection}}
    \underbrace{\mathbf{x}_s \in \mathcal{V}_{\text{sem}}^K}_{\text{semantic tokens}}
\end{equation}
The unified space not only facilitates direct cross-modal generation but also enables flexible compositional reasoning within a single architecture, going beyond the capabilities of conventional token or embedding alignment frameworks.

%创新点2:讲明白对交互感知的注意力构建的细节，以及应对这两个互逆任务的适配性与有效性；
\subsection{Interaction-Aware Attention Module}\label{sec:iaa}

%说明我们为什么要设计这样的一个模块；
To robustly capture the structured relationships underlying human-object interactions across both HOI detection (vision-to-semantics) and image generation (semantics-to-vision), we propose an Interaction-Aware Attention (IAA) module, which is designed to support both HOI detection and image generation tasks effectively. Prior approaches~\cite{InteractDiffusion,GLIGENOG} typically employ asymmetric architectures or separate attention modules for detection and generation, which restrict flexibility. In contrast, our IAA module adopts a parameter-shared, symmetric cross-attention mechanism that supports both tasks within a single structure, which enables consistent, context-aware alignment between vision and semantics.

\textbf{Design of Symmetric Cross-Attention.} 
IAA is implemented as a single, parameter-shared cross-attention block, whose directionality is determined solely by the task: in detection, \emph{visual tokens} acts as queries and \emph{semantic tokens} serves as keys and values; for generation, this assignment is reversed. Each token is enriched with a learnable \emph{modality-type embedding} $\mathbf{e}_t$, indicating whether it originates from the visual or semantic space, helping the model distinguish between heterogeneous sources during attention calculation.

Formally, let $\mathbf{Q}$, $\mathbf{K}$, and $\mathbf{V}$ be the query, key, and value matrices (the assignment depends on the task direction), and let $\mathbf{e}_q$, $\mathbf{e}_k$ be modality-type embeddings for queries and keys. The cross-modal attention in IAA is defined as:
\begin{equation}
    \mathbf{A}_{\text{cross}} = \mathrm{softmax}\left(
        \frac{(\mathbf{Q} + \mathbf{e}_q)(\mathbf{K} + \mathbf{e}_k)^\top}{\sqrt{d}}
    \right) \mathbf{V}
    \label{eq:symm_att}
\end{equation}
\begin{equation}
    \mathbf{H}_{\text{out}} = \mathbf{A}_{\text{cross}} + \mathbf{Q}
    \label{eq:att_output}
\end{equation}
Here, Eq.~\eqref{eq:symm_att} computes attention with awareness of each token's modality, and Eq.~\eqref{eq:att_output} applies a residual connection to preserve original query information.

%展示IAA在互逆任务中对交互语义的关注；双向箭头展示了视觉token和文本token之间的双向映射关系；同时从prompt-image-hoi-triplets的转换表现视觉和文本token在统一token space下不同任务中的相互转换关系
\begin{figure}[t]
  \centering
  \includegraphics[width=\linewidth]{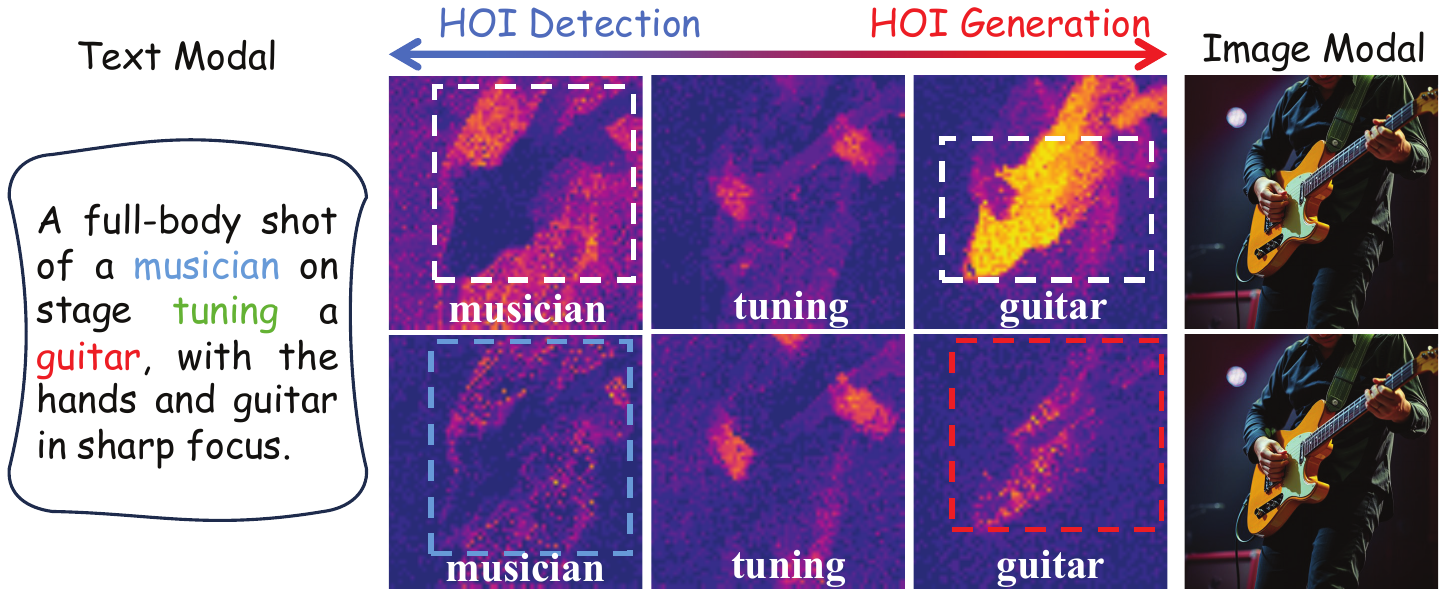}
   \caption{Visualization of interaction-aware attention maps produced by IAA for HOI detection and generation tasks. Bidirectional arrows indicate the mutual mapping between visual and textual tokens, highlighting IAA's capability in cross-modal interactive semantic modeling. The transitions among prompts, images, and HOI triplets further demonstrate unified token transformations across inverse tasks.}
   \label{fig:attention_map}
\end{figure}

%说明我们的注意力重点在结构对称，便于互逆任务的统一完成；所以没有设计细粒度标记（人，动作，物体）这种显式偏置引入，并证明粗粒度的标记也可以完成交互部分的关注；
\textbf{Adaptability and Effectiveness.}
Although we use only coarse-grained modality-type embeddings (``visual'' or ``semantic''), the IAA module exploits the structured HOI triplet input and the capacity of self-attention to implicitly align semantic slots with visual regions. Notably, our strictly symmetric design shares cross-attention architecture and parameters for both HOI detection and image generation, supporting flexible input-role swapping. As shown in Figure~\ref{fig:attention_map}, IAA accurately captures the spatial distribution of HOI triplets across both inverse tasks, enabling efficient and fine-grained cross-modal reasoning.

%创新点3:讲明白我们如何利用不同监督条件下的数据，借助unified token space完成统一的训练并用于两种互逆任务；
\subsection{Semi-supervised Learning Strategy}\label{sec:semi_supervised}

\textbf{Unified Cycle Consistency in Token Space.}
We leverage the unified token space to enable effective learning from fully, weakly, or unlabeled data, and to support both HOI detection and interaction-aware image generation as mutually inverse tasks. Unlike prior methods that rely on pseudo-labeling or disjoint modality-specific objectives, our approach introduces a dual cycle-consistency framework that enables joint training with heterogeneous supervision.

Specifically, let $\mathcal{F}$ and $\mathcal{G}$ denote the detection and generation modules, respectively; $\mathcal{T} \in \mathcal{V}_{\text{sem}}$ denotes semantic token sequences (\eg, HOI triplets), and $\mathcal{I} \in \mathcal{V}_{\text{code}}$ denotes visual tokens. We formulate the cross-modal cycle losses as:

\begin{equation}
\begin{aligned}
   \mathcal{L}_{\text{Cycle}}^{\text{sem} \rightarrow \text{code} \rightarrow \text{sem}} 
   &= \mathbb{E}_{\mathcal{T}}\big[
         \|\mathcal{F}(\mathcal{G}(\mathcal{T})) - \mathcal{T}\|_1
   \big] \\
   \mathcal{L}_{\text{Cycle}}^{\text{code} \rightarrow \text{sem} \rightarrow \text{code}} 
   &= \mathbb{E}_{\mathcal{I}}\big[
         \|\mathcal{G}(\mathcal{F}(\mathcal{I})) - \mathcal{I}\|_1
   \big]
\end{aligned}
\end{equation}

%讲清楚loss过程，就是用检测分支来检验生成，用生成分支来检验检测。
This dual cycle-consistency objective enforces that both semantic-to-visual-to-semantic and visual-to-semantic-to-visual mappings remain consistent in the unified token space. For each sample, the model either reconstructs semantics from generated visual tokens or reconstructs visuals from inferred semantic tokens. The cycle loss measures the difference between the initial and reconstructed token sequences in both directions. This unified training enables efficient use of data with any level of supervision, facilitating robust and effective learning of detection and generation tasks even with limited annotations.

%在这样的半监督框架下我们如何更好的设计训练过程；并讲明白我们的loss函数
\subsection{Model Training}\label{sec:model_training}

UniHOI is trained end-to-end with a semi-supervised paradigm that fully exploits the intrinsic duality between HOI detection and generation within a unified token space. Instead of optimizing detection and generative branches separately, all supervision—regardless of annotation granularity—is formulated as mutually inverse, cycle-consistent transformations and reconstructions in this unified space, enabling maximal alignment between original and reconstructed modalities.

Formally, for any input $x$ (image, HOI triplet, or interaction caption), we define the unified cycle mapping:
\begin{equation}
  x \xrightarrow{\mathcal{F}} y \xrightarrow{\mathcal{G}} \hat{x},\quad
  \mathcal{L}_{\text{cycle}} = d(x, \hat{x}),
\end{equation}
where $\mathcal{F}$ and $\mathcal{G}$ are invertible, parameter-shared detection and generation functions, $d(\cdot)$ denotes the sequence-level cross-entropy loss between the original and reconstructed tokens, with all tokens residing in the same unified space.

This framework naturally accommodates heterogeneous data: for paired data, both directions are optimized to enforce alignment; for single-modality or unlabeled data, available information is propagated through the cycle for reconstruction, obviating the need for handcrafted pseudo-labels or task-specific losses.

\noindent \textbf{Unified Loss Formulation.}
The overall loss aggregates all cycle pathways with dynamic weights, together with semantic alignment and diversity objectives:
\begin{equation}
\mathcal{L}_{\text{uni}} = \sum_{i=1}^N \lambda^{(i)} \mathcal{L}_{\text{cycle}}^{(i)} + \mathcal{L}_{\text{sem\_align}} + \mathcal{L}_{\text{div}},
\end{equation}
where $N$ is the number of cycle pathways (\eg, for paired, unpaired, or weakly supervised samples), $\mathcal{L}_{\text{cycle}}^{(i)}$ is the cycle-consistency loss for the $i$-th data type, and the weighting coefficients $\lambda^{(i)}$ are set to $\lambda^{(i)}=1$. $\mathcal{L}_{\text{sem\_align}}$ encourages semantic and spatial consistency, while $\mathcal{L}_{\text{div}}$ promotes generation diversity. Both are computed from attention maps to leverage spatial and statistical properties during training.

\begin{table*}[htbp]
  \centering
  \resizebox{\textwidth}{!}{
  \begin{tabular}{l | c | ccc | ccc | cc}
    \toprule
    \multirow{3}{*}{\textbf{Method}} & \multirow{3}{*}{\textbf{Backbone}}& \multicolumn{6}{c|}{\textbf{HICO-DET}} & \multicolumn{2}{c}{\textbf{V-COCO}} \\
    &&\multicolumn{3}{c}{Default} & \multicolumn{3}{c|}{Known Object} & \multicolumn{2}{c}{} \\
    \cmidrule(lr){3-5} \cmidrule(lr){6-8} \cmidrule(lr){9-10}
     & & Full & Rare & Non-rare & Full & Rare & Non-rare & $AP^{S1}_{role}$ & $AP^{S2}_{role}$  \\
    \midrule
    BCOM~\cite{hoi_det1}    & R50+CLIP & 39.34 & 39.90 & 39.17 & 42.24 & 42.86 & 42.05 & 65.10 & 69.90  \\
    MP-HOI~\cite{hoi_det2} & Swin-L & 44.53 & 44.48 & 44.55 & - & - & - & 66.22 & 67.64  \\
    SICHOI~\cite{hoi_det3}       & R101+ViT-L/16 & 45.04 & 45.61 & 44.88 & 48.16 & 48.37 & 48.09 & 71.13 &  75.62 \\
    PA-HOI~\cite{hoi_det4}        & Swin-L &   46.01    &   46.74    &   45.80   &   49.50    &   50.59    &   49.18    & 63.04 & 68.75  \\
    HOI-IDiff~\cite{hoi_det5}     & Diffusion &   47.71    &   48.36    &   47.52    &   50.56    &   51.95    &   50.14    & \textbf{73.42} & 76.13  \\
    % \midrule
    \rowcolor{row_color}\textbf{UniHOI (Ours)}         & VQGAN & 
    \textbf{48.16} & \textbf{50.74} & \textbf{48.12} & 
    \textbf{51.34} & \textbf{53.72} & \textbf{50.33} &
    72.91 &\textbf{77.45}  \\
    \bottomrule
  \end{tabular}
  }
  \caption{Comparison of state-of-the-art methods on HICO-DET (Default / Known Object) and V-COCO benchmarks (mAP scores).
    Best results are highlighted in bold. $AP_{\text{role}}^{S1}$ and $AP_{\text{role}}^{S2}$ denote standard splits on V-COCO.}
  \label{tab:hoi_det}
\end{table*} 

\begin{table*}[t]
\centering
\resizebox{\textwidth}{!}{
\begin{tabular}{lccccccccc}
\toprule
\multirow{2}{*}{\textbf{Method}} 
  & \multirow{2}{*}{\makecell[c]{\textbf{Wise}\\
  \textbf{Score}$\uparrow$}} 
  & \multirow{2}{*}{\makecell[c]{\textbf{Image}\\ \textbf{Reward}$\uparrow$}} 
  & \multirow{2}{*}{\makecell[c]{\textbf{FID}\\
  \textbf{Score}$\downarrow$}}
  & \multirow{2}{*}{\makecell[c]{\textbf{CLIP}\\ \textbf{Score}$\uparrow$}} 
  & \multirow{2}{*}{\makecell[c]{\textbf{HOI}\\ \textbf{Score}$\uparrow$}}
  & \multirow{2}{*}{\makecell[c]{\textbf{Interaction}\\
  \textbf{Accuracy}$\uparrow$}}
  & \multicolumn{3}{c}{\textbf{GenAIEval}$\uparrow$}\\
\cmidrule(lr){8-10}
  & & & & &  & & \textbf{Single Obj.} & \textbf{Two Obj.} & \textbf{Position} \\
\multicolumn{10}{c}{\cellcolor{gray!20}$\blacktriangledown$~\textit{Generation Only}} \\
InteractDiffusion~\cite{InteractDiffusion}    & -    & 0.79    & 38.2    & 13.43    & 0.40    & 0.22    & 0.71    & 0.34    & 0.07    \\
DALL-E2~\cite{dalle2}     & -    & 0.83    & 28.6    & 25.20     & 0.48    & 0.29    & 0.94    & 0.66    & 0.10    \\
SDXL~\cite{sdxl}       & 0.43 & 1.13 & 19.1 & 30.87 & 0.54 & 0.38 & 0.98 & 0.74 & 0.15 \\
SDv3.5~\cite{sd3}       & \textbf{0.51}    & 1.15 & \textbf{17.7}    & 31.54    & 0.56    & 0.35    & 0.96    & 0.71    & 0.14    \\
\multicolumn{10}{c}{\cellcolor{gray!20}$\blacktriangledown$~\textit{Unified Understanding \& Generation}} \\
Chameleon~\cite{chameleon}   & -    & 0.83    & 27.3    & 20.32    & 0.41    & 0.28    & -    & -    & -    \\
Show-o~\cite{showo} & 0.28 & 0.92 & 24.7 & 28.94 & 0.46 & 0.31 & 0.95 & 0.52 & 0.11 \\
Janus~\cite{janus}  & 0.16 & 1.03 & 22.1 & 29.45 & 0.50 & 0.36 & 0.97 & 0.68 & 0.28 \\
Liquid~\cite{liquid}     & -    & -    & 25.8    & 21.73    & 0.39    & 0.26    & -    & -    & -    \\
VAR-GPT~\cite{vargpt} & -   & 0.94 & 23.8 & 28.85 & 0.44 & 0.33 & 0.96 & 0.53 & 0.13 \\
\rowcolor{row_color}\textbf{UniHOI (Ours)} & 0.50 & \textbf{1.17} & 18.2 & \textbf{32.46} & \textbf{0.64} & \textbf{0.54} & \textbf{0.99} & \textbf{0.76} & \textbf{0.42} \\
\bottomrule
\end{tabular}}
\caption{
Evaluation results on HOI-oriented image generation benchmarks. We report Wise Score, Image Reward, FID, and CLIP Score to assess perceptual quality and text-image correspondence; HOI Score and Interaction Accuracy (IA) for the correctness of human-object interactions; and GenAIEval submetrics (SingleObj, TwoObj, Position) for compositional and spatial understanding. Lower FID and higher values for all other metrics indicate better performance. 
}
\label{tab:hoi_generation}
\end{table*}

\section{Experiments}
\label{sec:experiments}

\subsection{Datasets}
\label{sec:datasets}
%介绍我们的数据集组成以及特点，具有open-world hoi理解的能力；
UniHOI is trained semi-supervisedly using a curated mixture of three data types to balance supervision and scalability: \textbf{10\%} strongly-supervised, \textbf{25\%} weakly-supervised, and \textbf{65\%} unsupervised data. Compared to HICO-DET, our dataset not only covers nearly all its interaction types, but also contains a substantially larger and more diverse set of HOI triplets mined from open-domain sources, which facilitates truly open-world HOI understanding.

\textbf{Strongly-supervised:} We use HICO-DET~\cite{hico-det} and V-COCO~\cite{V-COCO}, which provide detailed HOI triplet annotations with bounding boxes and role labels, supplying rich spatial-semantic supervision.

\textbf{Weakly-supervised:} 150K image-text pairs from LAION-SG~\cite{laion-sg} are used, where triplets extracted from scene graphs serve as weak HOI signals, alleviating the long-tail annotation problem.

\textbf{Unsupervised:} 400K image-text pairs from LAION-400M~\cite{LAION-400M} are employed, utilizing only the captions for contrastive vision-language learning, without any structured labels.

%这里主要要说明HOI score 和IA是如何计算的，以及其合理性；
\subsection{Evaluation Metrics}
\label{sec:metrics}

% We evaluate UniHOI on both detection and generation tasks using standard and tailored metrics:

\textbf{HOI Detection:} We report mean Average Precision (mAP) on HICO-DET~\cite{hico-det} and V-COCO~\cite{V-COCO}, focusing on interaction instance mAP under both Default and Rare settings.

\noindent \textbf{HOI Generation:} We assess generation quality using 10K prompts from
our test set to compute Wise Score~\cite{wise}, Image Reward~\cite{imagereward}, FID, and CLIP Score~\cite{clip}. For HOI-specific evaluation, we report \textbf{HOI Score}, which measures the accuracy of the detected triplet against the input triplet using a pretrained HOI detector~\cite{fgahoi}, and \textbf{Interaction Accuracy}, which assesses how well the generated interaction details match the prompt by comparing a caption derived from the detector's output with the input prompt. Additionally, we include GenEval~\cite{genai-bench} for comprehensive assessment.

%这里需要讲清楚我们的实验设置以及半监督策略的一些工程实现
\subsection{Implementation Details}

In our experiments, we employ Llama3-8B~\cite{llama3} as the backbone of UniHOI, and adopt VQGAN~\cite{vqgan} from Chameleon~\cite{chameleon} as the image tokenizer. 
Model finetuning is performed using the Adam optimizer with a constant learning rate of $5\times10^{-4}$, following a linear warm-up over the first 10{,}000 iterations. 
UniHOI is trained for 700{,}000 iterations in total, with a per-device batch size of 8 on 32 NVIDIA H800 GPUs; to further increase the effective batch size to 16, we employ gradient accumulation with a step size of 2. A temperature parameter of $\tau=0.07$ is consistently used across all experiments. To mitigate the potential inefficiency arising from semi-supervised learning, we utilize a hybrid data loading strategy that ensures balanced sampling from both labeled and unlabeled datasets within each mini-batch.
 
\begin{figure*}[t]
  \centering
  \includegraphics[width=\textwidth]{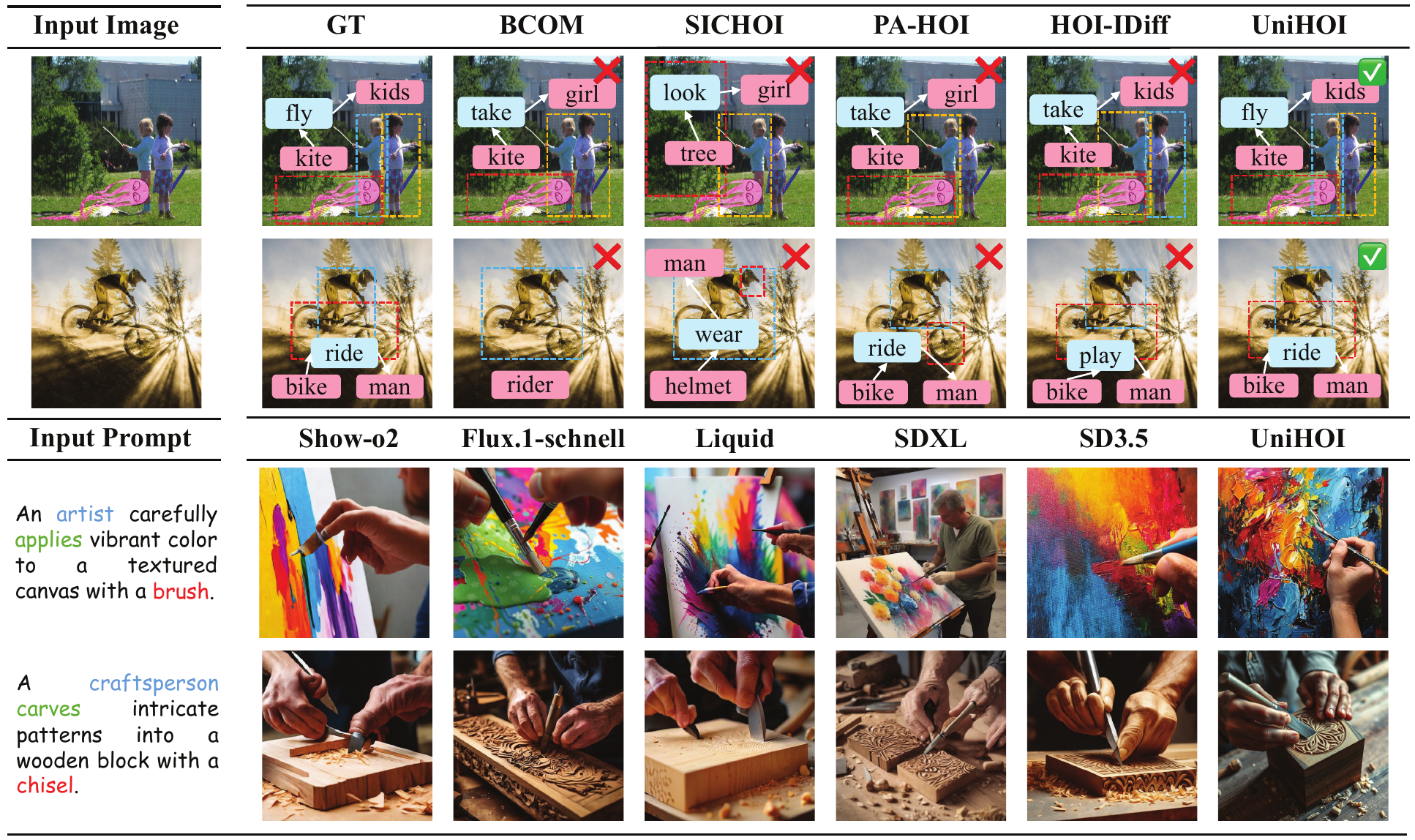}
   \caption{Qualitative results of UniHOI. For HOI detection, UniHOI demonstrates enhanced fine-grained interaction understanding; for HOI generation, it produces detailed interactive scenes, including realistic hand poses and precise tool usage.}
   \label{fig:qualitative}
\end{figure*}

\subsection{Quantitative \& Qualitative Results}
%通过数据，讲明白我们方法的优异性能以及为什么会在hoi detection上有效；要提及长为任务的表现；
\noindent \textbf{Quantitative Results on HOI Detection.}
Table~\ref{tab:hoi_det} presents a comprehensive comparison with state-of-the-art methods on HICO-DET and V-COCO benchmarks. UniHOI establishes new state-of-the-art results on HICO-DET, achieving 48.16 mAP on the full test set, 50.74 mAP on rare categories, and 51.34 mAP under the Known Object setting. On V-COCO, UniHOI achieves 72.91 in $AP_{\text{role}}^{S1}$ and the highest $AP_{\text{role}}^{S2}$ score of 77.45, demonstrating clear superiority in precise role-object localization despite a slightly lower $AP_{\text{role}}^{S1}$ compared to HOI-IDiff~\cite{hoi_det5}. These consistent improvements, particularly on rare and long-tailed HOI categories, further demonstrating the mutual benefits of jointly modeling HOI detection and generation.

\noindent \textbf{Quantitative Results on HOI Generation.}
%结合数据说明我们方法在hoi generation上的性能以及哪个模块最有贡献？
Table~\ref{tab:hoi_generation} presents a comprehensive quantitative comparison on HOI-oriented image generation benchmarks. UniHOI consistently achieves superior results across all major metrics, including the best Image Reward (1.17), lowest FID score (18.2), and highest CLIP Score (32.46), highlighting its strong perceptual quality and text-image correspondence. UniHOI also establishes new state-of-the-art performance on all GenAIEval submetrics, demonstrating excellent capability in compositional and spatial understanding. In terms of interaction-focused metrics such as HOI Score and Interaction Accuracy, our method significantly outperforms existing approaches, surpassing the previous state-of-the-art by 14.3\% and 42.0\%, respectively. This highlights the strong capability of our approach in generating images with fine-grained interactions. These results collectively validate the effectiveness of our unified framework in generating high-quality images that accurately represent complex human-object interactions.

\noindent \textbf{Qualitative Results.} 
%通过定型实验简要分析我们的方法在hoi detection上对于hardcase展现的优异性能，尽量往open-world能力上靠；hoi generation 对比比较明显，说明我们模型在交互细节部分生成能力优异；
Shown in Figure~\ref{fig:qualitative}, compared to state-of-the-art models in HOI detection, our method demonstrates superior fine-grained perception when handling hard cases. By leveraging open-world knowledge acquired under a unified label space, our approach extracts more accurate HOI triplets. For HOI generation, our method exhibits clear advantages in generating detailed and natural interactions compared to existing models—for instance, enabling more natural tool usage, such as picking up a paintbrush, with highly faithful hand pose information.

\begin{table}[htbp]
  \centering
  \resizebox{\columnwidth}{!}{
  \begin{tabular}{l|cc|cc}
    \toprule
    \multirow{2}{*}{\textbf{Method}} 
    & \multicolumn{2}{c|}{\textbf{HICO-DET (Default) $\uparrow$}} 
    & \multicolumn{2}{c}{\textbf{Generation $\uparrow$}} \\
    & Full & Rare & HOI Score & IA \\
    \midrule
    Separate Emb.         & 47.03 & 48.38 & 0.44 & 0.32 \\
    Shared Emb. Only      & 47.48 & 48.92 & 0.52 & 0.41 \\
    w/o Type Emb.         & 47.62 & 49.13 & 0.57 & 0.46 \\
    \textbf{Ours (Full)}  & \textbf{48.16} & \textbf{50.74} & \textbf{0.64} & \textbf{0.54} \\
    \bottomrule
  \end{tabular}
  }
\caption{
Ablation on \textbf{unified token space}. ``Separate Emb." uses unshared visual and text embeddings; ``Shared Emb. Only" shares embeddings but not vocabularies; ``w/o Type Emb." unifies vocab and embeddings but removes type information; 
}
\label{tab:ablation_unified}
\end{table}

\subsection{Ablation Study}

%unified token space的消融
\textbf{Ablation on Unified Token Space.} As shown in Table~\ref{tab:ablation_unified}, models without modality-type embedding or with separated embeddings achieve consistently lower performance in both detection (Full mAP: 48.16\,→\,47.62/47.03) and generation (HOI Score: 0.64\,→\,0.57/0.44), verifying that our unified, modality-aware token space substantially enhances both tasks by enabling more effective information fusion across modalities.

%IAA的消融
\noindent \textbf{Ablation on Interaction-Aware Attention.} As presented in Table~\ref{tab:ablation_iaa}, substituting IAA with a standard cross-attention module or removing modality-type embedding consistently degrades performance across all metrics, especially on rare HOIs and interaction accuracy (\eg, IA: 0.54\,→\,0.49/0.39). This demonstrates that our symmetric, modality-aware attention design is crucial for effective and generalizable cross-modal modeling. More ablation studies are provided in the supplementary material.

\begin{table}[htbp]
  \centering
  \resizebox{\columnwidth}{!}{
  \begin{tabular}{l|cc|cc}
    \toprule
    \multirow{2}{*}{\textbf{Setting}} 
    & \multicolumn{2}{c|}{\textbf{HICO-DET (Default) $\uparrow$}} 
    & \multicolumn{2}{c}{\textbf{Generation $\uparrow$}} \\
    & Full & Rare & HOI Score & IA \\
    \midrule
    w/o IAA            & 47.10 & 49.03 & 0.51 & 0.39 \\
    w/o Type Emb.          & 47.71 & 49.55 & 0.59 & 0.49 \\
    \textbf{Ours(Full)} & \textbf{48.16} & \textbf{50.74} & \textbf{0.64} & \textbf{0.54} \\
    \bottomrule
  \end{tabular}
  }
  \caption{
    Ablation on \textbf{interaction-aware attention}. ``w/o IAA'' denotes removing interaction-aware attention; ``w/o Type Emb.'' removes modality-type embedding from the symmetric module. Generation columns report HOI Score and Interaction Accuracy.
  }
  \label{tab:ablation_iaa}
\end{table}

\noindent \textbf{Ablation on Supervision Ratio.}
We further examine the cross-task benefits between HOI detection and generation under varying supervision ratios. Table~\ref{tab:ablation_data_ratio} demonstrates that increasing the proportions of weakly and unsupervised data boosts performance on both HOI detection (HOI Det.) and HOI generation (HOI Gen.). These results highlight a mutual promotion between detection and generation tasks, where supervision from one task benefits the other, proving the effectiveness of our multi-task and semi-supervised learning strategy.

\begin{table}[ht]
\centering
\resizebox{\linewidth}{!}{%
\begin{tabular}{ccc|ccc}
\toprule
\textbf{Strong} & \textbf{Weak} & \textbf{Unsupervised} & \textbf{HOI Det.} & \textbf{HOI Gen.} \\
\midrule
100\% & --    & --    & 45.62 & 0.22 \\
50\%  & 20\%  & 30\%  & 46.43 & 0.38 \\
20\%  & 20\%  & 60\%  & 47.65 & 0.59 \\
10\%  & 25\%  & 65\%  & \textbf{48.16} & \textbf{0.64} \\
\bottomrule
\end{tabular}
}
\caption{Ablation on \textbf{supervision ratios}. Varying the proportion of strong, weak, and unsupervised data shows the effect on HOI detection and generation. HOI Det. denotes the Full metric under the Default category in HICO-DET, and HOI Gen. denotes the HOI Score.}
\label{tab:ablation_data_ratio}
\end{table}

\section{Conclusion}
\label{sec:conclusion}

In this paper, we present UniHOI, a unified semi-supervised multimodal framework that jointly addresses human-object interaction (HOI) detection and generation via a modality-aware token space and symmetric interaction-aware attention modules. By unifying visual and semantic representations at the token level and leveraging a symmetric cross-modal attention mechanism, UniHOI achieves robust and generalizable performance for both tasks, substantially advancing the state of the art on benchmark datasets. Our approach enables effective knowledge and data sharing across HOI detection and generation, thereby improving data efficiency and long-tail generalization in open-world scenarios. Extensive experiments demonstrate that unified tokenization is crucial for flexible, compositional, and accurate cross-modal reasoning. We believe UniHOI not only advances the unified modeling of inverse HOI tasks, but also provides new insights for bridging recognition and generation in broader multimodal and open-vocabulary contexts.

\section{Acknowledgments}
This work was supported by the STI 2030-Major Projects under Grant No. 2022ZD0208801 and the NSFC under grant No. 62088102.

\bibliography{aaai2026}

@String(PAMI = {IEEE Trans. Pattern Anal. Mach. Intell.})

@String(CVPR= {IEEE Conf. Comput. Vis. Pattern Recog.})

@String(ICCV= {Int. Conf. Comput. Vis.})

@String(ECCV= {Eur. Conf. Comput. Vis.})

@String(NIPS= {Adv. Neural Inform. Process. Syst.})

@String(ACMMM= {ACM Int. Conf. Multimedia})

@String(ICLR = {Int. Conf. Learn. Represent.})

@String(PAMI  = {IEEE TPAMI})

@String(CVPR  = {CVPR})

@String(ICCV  = {ICCV})

@String(ECCV  = {ECCV})

@String(NIPS  = {NeurIPS})

@String(ACMMM = {ACM MM})

@String(ICLR  = {ICLR})

@String(WACV = {WACV})

@String(ICML  = {ICML})

@String(JMLR  = {JMLR})

@String(ACL  = {ACL})

@inproceedings{GLIGENOG,
  title={GLIGEN: Open-Set Grounded Text-to-Image Generation},
  author={Yuheng Li and Haotian Liu and Qingyang Wu and Fangzhou Mu and Jianwei Yang and Jianfeng Gao and Chunyuan Li and Yong Jae Lee},
  booktitle = CVPR,
  year={2023},
}

@inproceedings{InteractDiffusion,
    author    = {Hoe, Jiun Tian and Jiang, Xudong and Chan, Chee Seng and Tan, Yap-Peng and Hu, Weipeng},
    title     = {InteractDiffusion: Interaction Control in Text-to-Image Diffusion Models},
    booktitle = CVPR,
    year      = {2024}
}

@inproceedings{freecontrol,
  title={FreeControl: Training-Free Spatial Control of Any Text-to-Image Diffusion Model with Any Condition},
  author={Mo, Sicheng and Mu, Fangzhou and Lin, Kuan Heng and Liu, Yanli and Guan, Bochen and Li, Yin and Zhou, Bolei},
  booktitle = CVPR,
  year      = {2024}
}

@inproceedings{densediffusion,
  title={Dense Text-to-Image Generation with Attention Modulation},
  author={Kim, Yunji and Lee, Jiyoung and Kim, Jin-Hwa and Ha, Jung-Woo and Zhu, Jun-Yan},
  year={2023},
  booktitle = ICCV
}

@inproceedings{hico-det,
  title={Learning to Detect Human-Object Interactions},
  author={Yu-Wei Chao and Yunfan Liu and Michael Xieyang Liu and Huayi Zeng and Jia Deng},
  booktitle = WACV,
  year={2017},
}

@inproceedings{V-COCO,
  title={Visual Compositional Learning for Human-Object Interaction Detection},
  author={Zhi Hou and Xiaojiang Peng and Yu Qiao and Dacheng Tao},
  booktitle= ECCV,
  year={2020},
}

@inproceedings{LAION-400M,
  title={LAION-400M: Open Dataset of CLIP-Filtered 400 Million Image-Text Pairs},
  author={Christoph Schuhmann and Richard Vencu and Romain Beaumont and Robert Kaczmarczyk and Clayton Mullis and Aarush Katta and Theo Coombes and Jenia Jitsev and Aran Komatsuzaki},
  booktitle={arXiv preprint arxiv:2111.02114},
  year=2021,
}

@inproceedings{GLIDE,
  title={GLIDE: Towards Photorealistic Image Generation and Editing with Text-Guided Diffusion Models},
  author={Alex Nichol and Prafulla Dhariwal and Aditya Ramesh and Pranav Shyam and Pamela Mishkin and Bob McGrew and Ilya Sutskever and Mark Chen},
  booktitle=ICML,
  year={2021},
}

@inproceedings{dalle2,
  title={Hierarchical Text-Conditional Image Generation with CLIP Latents},
  author={Aditya Ramesh and Prafulla Dhariwal and Alex Nichol and Casey Chu and Mark Chen},
  year={2022},
  booktitle={arXiv preprint arxiv:2204.06125},
}

@inproceedings{stablediffusion,
  title={High-Resolution Image Synthesis with Latent Diffusion Models},
  author={Robin Rombach and A. Blattmann and Dominik Lorenz and Patrick Esser and Bj{\"o}rn Ommer},
  booktitle=CVPR,
  year={2022},
}

@inproceedings{Imagen,
  title={Photorealistic Text-to-Image Diffusion Models with Deep Language Understanding},
  author={Chitwan Saharia and William Chan and Saurabh Saxena and Lala Li and Jay Whang and Emily L. Denton and Seyed Kamyar Seyed Ghasemipour and Burcu Karagol Ayan and Seyedeh Sara Mahdavi and Raphael Gontijo Lopes and Tim Salimans and Jonathan Ho and David J. Fleet and Mohammad Norouzi},
  booktitle=NIPS,
  year={2022},
}

@inproceedings{clip,
  title={Learning Transferable Visual Models From Natural Language Supervision},
  author={Alec Radford and Jong Wook Kim and Chris Hallacy and Aditya Ramesh and Gabriel Goh and Sandhini Agarwal and Girish Sastry and Amanda Askell and Pamela Mishkin and Jack Clark and Gretchen Krueger and Ilya Sutskever},
  booktitle=ICML,
  year={2021},
}

@article{t5,
  title={Exploring the Limits of Transfer Learning with a Unified Text-to-Text Transformer},
  author={Colin Raffel and Noam M. Shazeer and Adam Roberts and Katherine Lee and Sharan Narang and Michael Matena and Yanqi Zhou and Wei Li and Peter J. Liu},
  journal=JMLR,
  year={2020},
}

@inproceedings{sdxl,
  title={SDXL: Improving Latent Diffusion Models for High-Resolution Image Synthesis},
  author={Dustin Podell and Zion English and Kyle Lacey and A. Blattmann and Tim Dockhorn and Jonas Muller and Joe Penna and Robin Rombach},
  booktitle=ICLR,
  year={2024},
}

@inproceedings{p2p,
  title={Null-text Inversion for Editing Real Images using Guided Diffusion Models},
  author={Mokady, Ron and Hertz, Amir and Aberman, Kfir and Pritch, Yael and Cohen-Or, Daniel},
  booktitle={arXiv preprint arXiv:2211.09794},
  year={2022}
}

@inproceedings{p2p-hrv,
  title={Cross-Attention Head Position Patterns Can Align with Human Visual Concepts in Text-to-Image Generative Models},
  author={Park, Jungwon and Ko, Jungmin and Byun, Dongnam and Suh, Jangwon and Rhee, Wonjong},
  booktitle= ICLR,
  year={2025}
}

@inproceedinges{hotr,
  title={HOTR: End-to-End Human-Object Interaction Detection with Transformers},
  author={Bumsoo Kim and Junhyun Lee and Jaewoo Kang and Eun-Sol Kim and Hyunwoo J. Kim},
  booktitle = CVPR,
  year={2021},
}

@article{fgahoi,
  title={FGAHOI: Fine-Grained Anchors for Human-Object Interaction Detection},
  author={Shuailei Ma and Yuefeng Wang and Shanze Wang and Ying Wei},
  journal=PAMI,
  year={2023},
  pages={2415-2429},
}

@inproceedings{thid,
author = {Wang, Suchen and Duan, Yueqi and Ding, Henghui and Tan, Yap-Peng and Yap, Kim-Hui and Yuan, Junsong},
title = {Learning Transferable Human-Object Interaction Detectors with Natural Language Supervision},
booktitle = CVPR,
year      = {2022},
}

@inproceedings{RLIP,
  title={RLIP: Relational Language-Image Pre-training for Human-Object Interaction Detection},
  author={Hangjie Yuan and Jianwen Jiang and Samuel Albanie and Tao Feng and Ziyuan Huang and Dong Ni and Mingqian Tang},
  booktitle = NIPS,
  year      = {2022}
}

@inproceedings{interactgan,
  title={InteractGAN: Learning to Generate Human-Object Interaction},
  author={Chen Gao and Si Liu and Defa Zhu and QUAN LIU and Jie Cao and Haoqian He and Ran He and Shuicheng Yan},
  booktitle = ACMMM,
  year={2020},
}

@inproceedings{score,
  title={Score-Based Generative Modeling through Stochastic Differential Equations},
  author={Yang Song and Jascha Narain Sohl-Dickstein and Diederik P. Kingma and Abhishek Kumar and Stefano Ermon and Ben Poole},
  booktitle = ICLR,
  year={2021},
}

@inproceedings{FID,
  title={GANs Trained by a Two Time-Scale Update Rule Converge to a Nash Equilibrium},
  author={Martin Heusel and Hubert Ramsauer and Thomas Unterthiner and Bernhard Nessler and G{\"u}nter Klambauer and Sepp Hochreiter},
  booktitle=ACL,
  year={2017},
}

@article{liquid,
  title={Liquid: Language models are scalable multi-modal generators},
  author={Wu, Junfeng and Jiang, Yi and Ma, Chuofan and Liu, Yuliang and Zhao, Hengshuang and Yuan, Zehuan and Bai, Song and Bai, Xiang},
  journal={arXiv preprint arXiv:2412.04332},
  year={2024}
}

@misc{mmada,
      title={MMaDA: Multimodal Large Diffusion Language Models}, 
      author={Ling Yang and Ye Tian and Bowen Li and Xinchen Zhang and Ke Shen and Yunhai Tong and Mengdi Wang},
      year={2025},
      eprint={2505.15809},
      archivePrefix={arXiv},
      primaryClass={cs.CV},
      url={https://arxiv.org/abs/2505.15809}, 
}

@misc{vqgan,
      title={Taming Transformers for High-Resolution Image Synthesis}, 
      author={Patrick Esser},
      year={2020},
      eprint={2012.09841},
      archivePrefix={arXiv},
      primaryClass={cs.CV}
}

@misc{laion-sg,
      title={LAION-SG: An Enhanced Large-Scale Dataset for Training Complex Image-Text Models with Structural Annotations}, 
      author={Zejian Li and Chenye Meng and Yize Li and Ling Yang and Shengyuan Zhang and Jiarui Ma and Jiayi Li and Guang Yang and Changyuan Yang and Zhiyuan Yang and Jinxiong Chang and Lingyun Sun},
      year={2024},
      eprint={2412.08580},
      archivePrefix={arXiv},
      primaryClass={cs.CV},
      url={https://arxiv.org/abs/2412.08580}, 
}

@misc{janus-pro,
      title={Janus-Pro: Unified Multimodal Understanding and Generation with Data and Model Scaling}, 
      author={Xiaokang Chen and Zhiyu Wu and Xingchao Liu and Zizheng Pan and Wen Liu and Zhenda Xie and Xingkai Yu and Chong Ruan},
      year={2025},
      eprint={2501.17811},
      archivePrefix={arXiv},
      primaryClass={cs.AI},
      url={https://arxiv.org/abs/2501.17811}, 
}

@article{showo,
  title={Show-o: One Single Transformer to Unify Multimodal Understanding and Generation},
  author={Xie, Jinheng and Mao, Weijia and Bai, Zechen and Zhang, David Junhao and Wang, Weihao and Lin, Kevin Qinghong and Gu, Yuchao and Chen, Zhijie and Yang, Zhenheng and Shou, Mike Zheng},
  journal={arXiv preprint arXiv:2408.12528},
  year={2024}
}

@article{genai-bench,
  title={Evaluating Text-to-Visual Generation with Image-to-Text Generation},
  author={Lin, Zhiqiu and Pathak, Deepak and Li, Baiqi and Li, Jiayao and Xia, Xide and Neubig, Graham and Zhang, Pengchuan and Ramanan, Deva},
  journal={arXiv preprint arXiv:2404.01291},
  year={2024}
}

@inproceedings{hoi_det1,
  title={Bilateral adaptation for human-object interaction detection with occlusion-robustness},
  author={Wang, Guangzhi and Guo, Yangyang and Xu, Ziwei and Kankanhalli, Mohan},
  booktitle={Proceedings of the IEEE/CVF Conference on Computer Vision and Pattern Recognition},
  pages={27970--27980},
  year={2024}
}

@inproceedings{hoi_det2,
  title={Open-world human-object interaction detection via multi-modal prompts},
  author={Yang, Jie and Li, Bingliang and Zeng, Ailing and Zhang, Lei and Zhang, Ruimao},
  booktitle={Proceedings of the IEEE/CVF Conference on Computer Vision and Pattern Recognition},
  pages={16954--16964},
  year={2024}
}

@inproceedings{hoi_det3,
  author={Luo, Jinguo and Ren, Weihong and Jiang, Weibo and Chen, Xi'ai and Wang, Qiang and Han, Zhi and Liu, Honghai},
  booktitle={Proceedings of the IEEE/CVF Conference on Computer Vision and Pattern Recognition},
  pages={28212--28222},
  year={2024}
}

@inproceedings{hoi_det4,
  title={Exploring Pose-Aware Human-Object Interaction via Hybrid Learning},
  author={Wu, Eastman ZY and Li, Yali and Wang, Yuan and Wang, Shengjin},
  booktitle={Proceedings of the IEEE/CVF Conference on Computer Vision and Pattern Recognition},
  pages={17815--17825},
  year={2024}
}

@inproceedings{hoi_det5,
  title={An Image-like Diffusion Method for Human-Object Interaction Detection},
  author={Hui, Xiaofei and Qu, Haoxuan and Rahmani, Hossein and Liu, Jun},
  booktitle={Proceedings of the Computer Vision and Pattern Recognition Conference},
  pages={14002--14012},
  year={2025}
}

@inproceedings{sd3,
  title={Scaling rectified flow transformers for high-resolution image synthesis},
  author={Esser, Patrick and Kulal, Sumith and Blattmann, Andreas and Entezari, Rahim and M{\"u}ller, Jonas and Saini, Harry and Levi, Yam and Lorenz, Dominik and Sauer, Axel and Boesel, Frederic and others},
  booktitle={Forty-first international conference on machine learning},
  year={2024}
}

@article{chameleon,
  title={Chameleon: Mixed-modal early-fusion foundation models},
  author={Team, Chameleon},
  journal={arXiv preprint arXiv:2405.09818},
  year={2024}
}

@inproceedings{janus,
  title={Janus: Decoupling visual encoding for unified multimodal understanding and generation},
  author={Wu, Chengyue and Chen, Xiaokang and Wu, Zhiyu and Ma, Yiyang and Liu, Xingchao and Pan, Zizheng and Liu, Wen and Xie, Zhenda and Yu, Xingkai and Ruan, Chong and others},
  booktitle={Proceedings of the Computer Vision and Pattern Recognition Conference},
  pages={12966--12977},
  year={2025}
}

@article{vargpt,
  title={VARGPT: Unified Understanding and Generation in a Visual Autoregressive Multimodal Large Language Model},
  author={Zhuang, Xianwei and Xie, Yuxin and Deng, Yufan and Liang, Liming and Ru, Jinghan and Yin, Yuguo and Zou, Yuexian},
  journal={arXiv preprint arXiv:2501.12327},
  year={2025}
}

@misc{llava,
      title={Visual Instruction Tuning}, 
      author={Liu, Haotian and Li, Chunyuan and Wu, Qingyang and Lee, Yong Jae},
      publisher={NeurIPS},
      year={2023},
}

@inproceedings{blip2,
  title={Blip-2: Bootstrapping language-image pre-training with frozen image encoders and large language models},
  author={Li, Junnan and Li, Dongxu and Savarese, Silvio and Hoi, Steven},
  booktitle={International conference on machine learning},
  pages={19730--19742},
  year={2023},
  organization={PMLR}
}

@article{llama3,
  title={The llama 3 herd of models},
  author={Grattafiori, Aaron and Dubey, Abhimanyu and Jauhri, Abhinav and Pandey, Abhinav and Kadian, Abhishek and Al-Dahle, Ahmad and Letman, Aiesha and Mathur, Akhil and Schelten, Alan and Vaughan, Alex and others},
  journal={arXiv preprint arXiv:2407.21783},
  year={2024}
}

@article{imagereward,
  title={Imagereward: Learning and evaluating human preferences for text-to-image generation},
  author={Xu, Jiazheng and Liu, Xiao and Wu, Yuchen and Tong, Yuxuan and Li, Qinkai and Ding, Ming and Tang, Jie and Dong, Yuxiao},
  journal={Advances in Neural Information Processing Systems},
  volume={36},
  pages={15903--15935},
  year={2023}
}

@article{wise,
  title={Wise: A world knowledge-informed semantic evaluation for text-to-image generation},
  author={Niu, Yuwei and Ning, Munan and Zheng, Mengren and Jin, Weiyang and Lin, Bin and Jin, Peng and Liao, Jiaqi and Feng, Chaoran and Ning, Kunpeng and Zhu, Bin and others},
  journal={arXiv preprint arXiv:2503.07265},
  year={2025}
}

\end{document}